\documentclass[journal]{IEEEtranTIE}
\usepackage{graphicx} 
\usepackage{cite}
\usepackage{picinpar}
\usepackage{amsmath}
\usepackage{url}
\usepackage{flushend}
\usepackage[latin1]{inputenc}
\usepackage{colortbl}
\usepackage{soul}
\usepackage{multirow}
\usepackage{pifont}
\usepackage{color}
\usepackage{alltt}
\usepackage[hidelinks]{hyperref}
\usepackage{enumerate}
\usepackage{siunitx}
\usepackage{breakurl}
\usepackage{epstopdf}
\usepackage{pbox}

\usepackage{amsfonts}
\usepackage{algorithmic}
\usepackage{array}
\usepackage[caption=false,font=normalsize,labelfont=sf,textfont=sf]{subfig}
\usepackage{stfloats}
\usepackage{textcomp}
\usepackage{verbatim}
\usepackage{xcolor}
\usepackage{amssymb}
\usepackage{balance}
\usepackage{booktabs}
\DeclareMathAlphabet\mathbfcal{OMS}{cmsy}{b}{n}
\usepackage[ruled,vlined,linesnumbered]{algorithm2e}
\usepackage{amsthm}
\newtheorem{Definition}{\hspace{1em} Definition} 

\begin{document}
\title{\Large TRUST-Planner: Topology-guided Robust Trajectory Planner for AAVs with Uncertain Obstacle Spatial-temporal Avoidance}

\author{
	\vskip 1em
	Junzhi Li, \emph{Graduate Student Member, IEEE},
	Jingliang Sun,
	Jianxin Zhong
	and Teng Long
	\thanks{
		This work was supported in part by the National Natural Science Foundation of China under Grants 52272360, 52372347, and in part by Beijing Nova Program 20250484886.
		({\it Corresponding authors: Teng Long, Jingliang Sun})
	}
	\thanks{
		The authors are with the School of Aerospace Engineering, Beijing Institute of Technology, Beijing 100081, China;
		and with the Key Laboratory of Dynamics and Control of Flight Vehicle, Ministry of Education, Beijing 100081.
		(E-mails: \href{mailto:junzhi\_lee@bit.edu.cn}{junzhi\_lee@bit.edu.cn};
		\href{mailto:jlsun@bit.edu.cn}{jlsun@bit.edu.cn};
		\href{mailto:3120235048@bit.edu.cn}{3120235048@bit.edu.cn};
		\href{mailto:tenglong@bit.edu.cn}{tenglong@bit.edu.cn});
	}%
	\thanks{
		This paper has a supplementary video provided by the authors available at \href{https://www.bilibili.com/video/BV1RJWqzqEMz/}{https://www.bilibili.com/video/BV1RJWqzqEMz/}.
	}%
	\thanks{
		\textcolor{red}{
			This paper has been accepted by IEEE Transactions on Industrial Electronics (TIE) for publication. The final version will be available online at \url{https://ieeexplore.ieee.org/} after publication.}
	}%
}

\maketitle

\begin{abstract}
	Despite extensive developments in motion planning of autonomous aerial vehicles (AAVs), existing frameworks face the challenges of local minima in complex dynamic environments, leading to increased collision risks.
	To address these challenges, we present TRUST-Planner, a topology-guided hierarchical planner for robust spatial-temporal obstacle avoidance.
	In the frontend, a dynamic enhanced visible probabilistic roadmap (DEV-PRM) is proposed to explore topological paths for global guidance rapidly.
	The backend utilizes a uniform terminal-free minimum control polynomial (UTF-MINCO) to enable efficient predictive obstacle avoidance and fast computation.
	Furthermore, an incremental multi-branch trajectory management framework is introduced to enable spatial-temporal topological decision-making, while efficiently leveraging historical information to reduce replanning time.
	Simulation results show that TRUST-Planner outperforms baseline competitors, achieving millisecond-level computation efficiency and high success rate in tested complex environments.
	Real-world experiments further validate the feasibility and practicality of the proposed method.
\end{abstract}


\markboth{}%
{}

\definecolor{limegreen}{rgb}{0.2, 0.8, 0.2}
\definecolor{forestgreen}{rgb}{0.13, 0.55, 0.13}
\definecolor{greenhtml}{rgb}{0.0, 0.5, 0.0}

\section{Introduction} \label{sec: 1}

\IEEEPARstart{C}{urrently}, autonomous aerial vehicles (AAVs) play an crucial role in widespread areas, e.g., transportation, search and rescue \cite{doornbos_drone_2024}.
With the deepening of these applications, AAVs are required to operate in increasingly complex environments, where dynamic obstacles, such as pedestrians, vehicles, and other AAVs, present growing risks of collision \cite{degroot_topologydriven_2025}.
Therefore, ensuring both efficiency and safety in uncertain and dynamic environments remains a significant challenge.

Motion planning navigates AAVs to traverse obstacles and reach their designated goals safely and quickly \cite{quan_survey_2020}.
In general, it can be divided as high-level path planning and low-level trajectory planning.
Path planning \cite{luo_research_2024} mainly searches for discrete geometric waypoints from start to goal.
However, it usually neglects detailed dynamics for better computation efficiency and global exploration, limiting the full utilization of AAVs' maneuverability.
On the contrary, trajectory planning \cite{zhou_survey_2025} considers practical differential dynamics to generate continuous, smooth, and feasible motions.
Its outputs can be accurately tracked and executed by the bottom flight controller, thus providing better capabilities of real-time response and obstacle avoidance.
However, generating dynamically precise trajectories incurs sensitivity of the initial guess and more computational burdens.
Consequently, the prevailing strategy is to integrate both by hierarchical planning frameworks, utilizing path planning for global guidance and trajectory generation for dynamic feasibility and real-time execution.

\begin{figure}[!t]
	\centering
	\includegraphics[width=3.3in, trim=1 1 1 1, clip]{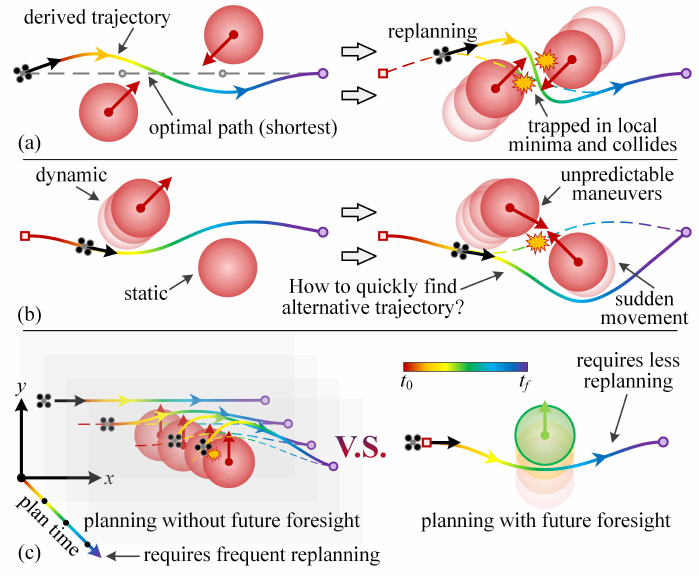}
	\caption{
		Main challenges of hierarchical planning in dynamic environments.
		(a) Structural spatial-temporal inconsistency.
		(b) Time-critical reconfiguration.
		(c) Lack of predictive foresight.
	}
	\label{fig: Challenges}
\end{figure}

However, as shown in Fig.\ref{fig: Challenges}, existing hierarchical frameworks face significant challenges in dynamic environments as:
\begin{enumerate}
	\item {\it Structural Spatial-temporal Inconsistency}.
	      Paths contain discrete geometric points without explicit time information, while trajectories involve strong coupling of time and space.
	      As a result of this inconsistency, although the guidance path is feasible or even global optimal, the derived trajectory may still be trapped in local minima and has high collision risks in dynamic environments.
	\item {\it Time-critical Reconfiguration}.
	      Dynamic obstacles may unpredictably and suddenly maneuver.
	      Without dedicated mechanisms for efficient and rapid reconfiguration in emergency handling, existing frameworks often fail to promptly find and switch to alternative trajectory, resulting in frequent abrupt stops and reduced flexibility.
	\item {\it Lack of Predictive Foresight}.
	      Existing frameworks often project dynamic planning on multiple stamps to simplify as static problems, and thus lack the predictive foresight of obstacle avoidance from a spatial-temporal perspective.
	      Thus, high-frequency replanning is often required for dynamic changes, imposing computational burdens.
\end{enumerate}

To address these issues, we propose a \textbf{T}opology-guided \textbf{R}obust Planner with \textbf{U}ncertain Obstacle \textbf{S}patial-\textbf{T}emporal Avoidance (TRUST-Planner).
Firstly, a dynamic enhanced visible PRM (DEV-PRM) frontend is proposed to capture spatial-temporal topological characteristics of dynamic environments.
Then, a uniform terminal-free minimum control polynomial (UTF-MINCO) backend is proposed for flexible dynamic obstacle avoidance with predictive foresight, which transcribes the trajectory planning into a lightweight unconstrained optimization for fast computation.
Furthermore, an incremental multi-branch trajectory management framework is introduced.
By incremental parallel replanning, it efficiently maintains the topological diversity of trajectories, enabling trajectory decision-making for breaking local minima.
Finally, extensive simulations and real-world experiments are presented to validate the efficiency and practicality of the proposed method.
The main contributions of this work are summarized as:
\begin{enumerate}
	\item An efficient DEV-PRM topological path searching frontend is proposed. Compared to the existing baseline \cite{zhou_robust_2020}, it improves the topological searching efficiency by 81.9\% for more diversified topological guidance.
	\item A lightweight trajectory planning backend UTF-MINCO is proposed for efficient dynamic obstacle avoidance with predictive foresight. Compared with \cite{long_Realtime_2023}, \cite{wang_geometrically_2022}, \cite{patterson_gpopsii_2014}, ours achieves 10~ms-level real-time optimization.
	\item An incremental topological trajectory management framework is introduced, which efficiently maintains the topological diversity using historical information and fast parallel optimization. In challenging, complex, dynamic, and unknown test environments, our framework outperforms baselines \cite{zhou_robust_2020}\cite{long_Realtime_2023} in terms of success rate.
\end{enumerate}

\section{Related Works} \label{sec: 2}

\subsection{Hierarchical Planning Framework}

Hierarchical planning has become a mainstream approach for motion planning in complex environments.
The core idea is to decouple complex planning problems into several sublevels \cite{lin_hierarchical_2014}, such as ``{\it{path$-$trajectory}}'' bi-level frameworks \cite{bry_aggressive_2015}, and ``{\it{path$-$corridor$-$trajectory}}'' tri-level frameworks \cite{wang_geometrically_2022}.
This decoupling improves the efficiency and alleviates the initial sensitivity of trajectory planning \cite{sun_safe_2025}.
Unfortunately, as shown in Fig.\ref{fig: Challenges}(a), it introduces a structural defect of spatial-temporal inconsistency, leading to local minima and potential collisions in dynamic environments.
Some researchers have attempted to introduce simplified kinematic \cite{jesus_MADER_2022}, or kinodynamic models \cite{zhou_Robust_2019}\cite{ding_Efficient_2019} into the path planning to provide coarse time information, while these methods often come at the cost of reduced computational efficiency.

Addressing these challenges, we focus on topology-guided hierarchical planning.
It aims to find multiple diverse homotopic solutions in complex environments \cite{liu_homotopy_2023a}.
Interestingly, the homotopy of paths or trajectories means that they lead to the same local results through continuous optimization \cite{bhattacharya_topological_2012a}.
Therefore, by searching the diversity of homotopic solutions, these methods show potential to break the local minima and achieve flexible reconstruction.
\cite{zhou_robust_2020}, \cite{ye_tgkplanner_2021}, \cite{sahin_topogeometrically_2025} demonstrated the efficiency and robustness of topology-guided planning for quadrotors operating in cluttered environments.
However, their approaches are developed for static scenarios and thus lack the foresight to handle spatial-temporal avoidance of moving obstacles.
\cite{degroot_topologydriven_2025}, \cite{yu_topologyguided_2024} explored topology-driven dynamic planning frameworks, but they are short of efficiently managing the known homotopies.
As a result, they may lose previously explored information during replanning, leading to reduced flexibility in trajectory reconstruction and potential collisions.

\subsection{Path Planning}

Path planning includes graph-based search \cite{lin_efficient_2024}, potential field \cite{pan_improved_2022a}, etc.
However, these methods typically focus on finding a single solution, lacking the capability to explore multiple homotopies.
By random sampling and multi-objective strategy, some sampling-based methods \cite{combelles_morrfx_2022} and intelligent optimization \cite{liu_cooperative_2024} can obtain Pareto frontiers of feasible paths.
Nevertheless, these multi-objective methods do not inherently guarantee the real-time solution of topologically distinct paths.

Currently, the probabilistic roadmap (PRM) \cite{schmitzberger_capture_2002} and its variants\cite{degroot_topologydriven_2025}\cite{zhou_robust_2020},  are the most widely used approaches for efficient topological path planning.
PRM-based methods typically have a simple but effective architecture, which builds a roadmap by fast customized sampling (e.g. visible graph \cite{zhou_robust_2020}, \cite{werner_approximating_2024}), and then uses homotopy checking (such as H-signature \cite{bhattacharya_topological_2012a}, visibility deformation (VD) \cite{zhou_robust_2020}, winding numbers\cite{kretzschmar_socially_2016}) to distinguish topological paths.
However, most of them are restricted in dynamic environments, as they do not account for the obstacles motions.
As a result, they can only capture spatial characteristics, limiting their ability to guide the spatial-temporal topological trajectories in dynamic environments.

\subsection{Trajectory Optimization}

Trajectory planning, also known as trajectory optimization, is typically formulated as an optimal control problem (OCP), which aims to minimize flight time \cite{wang_minimumtime_2017} or control efforts \cite{mellinger_minimum_2011}) subject to dynamics, terminal conditions and
collision avoidance.
Such an OCP is generally a nonlinear and nonconvex problem, and thus directly solving it by general numerical optimization \cite{chai_solving_2021a}  is often time-consuming and intractable.

Owing to the differential flatness of AAVs' dynamics (e.g., quadrotors\cite{mellinger_minimum_2011}, fixed-wing\cite{li_differential_2025a}, VTOL\cite{tal_aerobatic_2023}), trajectory optimization can be transcribed into simple problems such as linear programming (LP) \cite{long_Realtime_2023}, quadratic programming (QP)\cite{mellinger_minimum_2011}, and unconstrained optimization\cite{wang_geometrically_2022} for fast and efficient computation.
The core of these differential flatness-based methods is lightweight trajectory representation, including polynomials\cite{mellinger_minimum_2011},
B\'{e}zier \cite{jesus_faster_2022}, and other customized splines \cite{li_differential_2025a}.
Among them, minimum control polynomials (MINCO) \cite{wang_geometrically_2022} is a flexible and efficient representation with spatial-temporal deformability.
However, MINCO relies on fixed terminal states, which limit its ability to handle uncertain or changing endpoints in dynamic environments.
Moreover, as pointed out in \cite{quan_robust_2023}, the nonuniform distribution of control points may also cause small obstacles to be missed.

\section{Methodology} \label{sec: 3}

\subsection{Framework} \label{sec: 3.1}

\begin{figure}[!tp]
	\centering
	\includegraphics[width=3.49in, trim=1 1 1 1, clip]{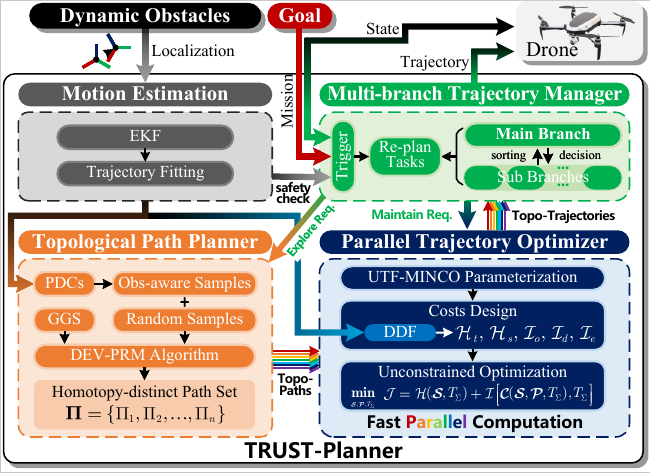}
	\caption{Framework of TRUST-Planner.}
	\label{fig: overall framework}
	\vspace{-0.1cm}
\end{figure}

As shown in Fig.\ref{fig: overall framework}, TRUST-Planner comprises three key modules that work collaboratively to enable effective and safe navigation for AAVs in complex dynamic environments. These modules are detailed in Secs.\ref{sec: 3.2}--\ref{sec: 3.4}.
\begin{enumerate}
	\item {\it Multi-branch Trajectory Manager}.
	      This module serves as a customized container for trajectory management.
	      Upon detecting environmental changes or task reconfiguration, the manager assigns replanning tasks to the path planner and trajectory optimizer, while continuously providing the optimal trajectory for drone navigation.
	\item {\it Topological Path Planner}.
	      We propose a Dynamic Enhanced Visible PRM (DEV-PRM) to capture approximate spatial-temporal topological characteristics in dynamic environments, which efficiently generates diverse, homotopy distinct paths for global guidance.
	\item {\it Parallel Trajectory Optimizer}.
	      The trajectory planning is formulated into a lightweight unconstrained optimization utilizing the proposed uniform terminal-free MINCO (UTF-MINCO) representation.
	      The optimizer takes the topological paths as initial guidance, and generates multiple feasible trajectories by fast parallel computation.
\end{enumerate}

\subsection{Multi-branch Topological Trajectory Management} \label{sec: 3.2}

Based on the concept of homotopic path \cite{bhattacharya_topological_2012a} and trajectory \cite{degroot_topologydriven_2025}, we first introduce the concept of the spatial-temporal homotopic trajectories as follows.
Unlike \cite{degroot_topologydriven_2025}, Definition 1 places greater emphasis on spatial-temporal characteristics, i.e., a trajectory may intersect with moving obstacles in space, but remains safe by avoiding them in both time and space.

\begin{Definition}[\it Spatial-temporal homotopic trajectories]
	For any two spatial-temporal trajectories $\tau_1(t)$, $\tau_2(t) \in \mathbb{R}^3 \times \mathbb{R} $ with $t \in [t_0, t_f]$, $\tau_1(t)$ is homotopic to $\tau_2(t)$ if there exists a continuous mapping $[t_0, t_f]\times[t_0, t_f]$.
	The trajectories in the same spatial-temporal homotopic class can be continuously deformed from one to another without crossing any obstacles at any time stamp $t \in [t_0, t_f]$.
\end{Definition}

\begin{figure}[!htp]
	\centering
	\includegraphics[width=3.49in, trim=1 1 1 1, clip]{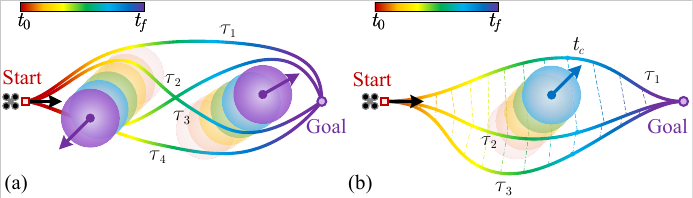}
	\caption{Illustration of spatial-temporal topologically trajectories.
		(a) Four trajectories $\tau_1$--$\tau_4$ belong to different spatial-temporal homotopy classes.
		(b) Temporal visibility deformation. $\tau_1$ and $\tau_2$ are in distinct homotopy because they are blocked by the moving obstacle at $t_c$, while $\tau_2$ and $\tau_3$ are equivalent.}
	\label{fig: spatial-temporal homotopy}
	\vspace{-0.5cm}
\end{figure}

\begin{algorithm}[!htp]
	\small
	\SetNlSty{}{}{}
	\SetAlgoNlRelativeSize{0}
	\SetFuncSty{textbf}    
	\SetFuncArgSty{text}     
	\caption{Incremental Planning of TRUST-Planner} \label{alg: trajectory manager}
	\SetKwProg{Fn}{Function}{:}{}
	\SetKwFunction{AddNewBranch}{AddNewBranch}
	\SetKwFunction{isTopoEqu}{isTopoEqu}
	\SetKwFunction{Sort}{Sort}
	\SetKwFunction{Size}{Size}
	\SetKwFunction{PopBack}{PopBack}
	\SetKwFunction{FwdMarch}{FwdMarch}
	\SetKwFunction{FwdExplore}{FwdExplore}
	\SetKwFunction{BwdReplan}{BwdReplan}
	\KwIn{Start $\mathbf{p}_{s}$, goal $\mathbf{p}_{g}$, dynamic obstacles $\mathcal{O}(t)$}
	$t \leftarrow 0$, $\mathbf{p}(t) \leftarrow \mathbf{p}_{s}$\;
	\While(\tcp*[f]{$\varepsilon$ is tolerance}){\rm $\|\mathbf{p}(t)-\mathbf{p}_{g}\| \geq \varepsilon$}{
	$t \leftarrow t + \Delta t$\;
	Update $\mathbf{p}(t)$, $\mathcal{O}(t)$\;
	\For{\rm each $\tau_i \in \mathbb{T}$}{
		\lIf{$i = 0$}
		{$\tau_0 \leftarrow$ \FwdMarch{$\tau_0$, $\mathbf{p}_g$, $\mathcal{O}(t)$}}
		\lElse
		{$\tau_i \leftarrow$ \BwdReplan{$\tau_i$, $\mathbf{p}(t)$, $\mathcal{O}(t)$}}
	}
	$\mathbb{T}_{\text{new}} \leftarrow$ \FwdExplore{$\tau_0$, $\mathbf{p}_g$, $\mathcal{O}(t)$}\;
	\lFor{\rm each $\tau_j \in \mathbb{T}_{\text{new}}$}{\AddNewBranch{$\mathbb{T}$, ${\tau}_{j}$}}
	}
	\vspace{-0.1cm}
	\hrule
	\vspace{0.05cm}
	\Fn{\AddNewBranch{$\mathbb{T}$, ${\tau}_{\text{new}}$}}{
	isTopoNew $\leftarrow$ true\;
	\For{\rm each $\tau_i \in \mathbb{T}$}{
	\uIf(\tcp*[f]{by TVD}){\rm \isTopoEqu{${\tau}_{i}$, ${\tau}_{\text{new}}$}}
	{
	\lIf{$\mathcal{J}_{\text{new}} < \mathcal{J}_i$ 
	}{
	${\tau}_{i}\leftarrow{\tau}_{\text{new}}$, isTopoNew=false}
	\lElse
	{\Return $\mathbb{T}$}
	}
	}
	\lIf(\tcp*[f]{sort by $\mathcal{J}$}){\rm isTopoNew}{$\mathbb{T} = \mathbb{T} \cup  \{{\tau}_{\text{new}}\}$}
	\lIf{\rm $\mathbb{T}$.\Size{} $>$ $N$}{$\mathbb{T}$.\PopBack{}}
	\Return $\mathbb{T}$\;
	\vspace{-0.2cm}
	}
\end{algorithm}

Inspired by the uniform visibility deformation (UVD) \cite{zhou_robust_2020} for determining spatial homotopy of paths, we extended this idea to the spatial-temporal domain by employing the temporal visibility deformation (TVD) as shown in Fig.\ref{fig: spatial-temporal homotopy},  to identify the spatial-temporal homotopy of trajectories in dynamic environments.
For two trajectories $\tau_1(t)$, $\tau_2(t)$, check whether they are blocked by the dynamic obstacles at sample the time stamps $t_c$.
They belong to the same homotopy if and only if
$\overrightarrow{\tau_1(t_c)\tau_2(t_c)} \cap \mathcal{O}(t_c) = \varnothing, \,\forall t_c \in [t_0, t_f]$.
By comparing the visibility at sample times, one can efficiently separate the trajectories into different spatial-temporal homotopy classes.

Utilizing the TVD for spatial-temporal homotopy checking, we propose an incremental planning framework to manage multi-branch trajectory branches and enable efficient trajectory replanning, as shown in Algorithm \ref{alg: trajectory manager}.
A customized queue with multiple different topological trajectory branches is maintained, denoted as $\mathbb{T} = \{ {\tau}_0, {\tau}_1,\ldots,{\tau}_{n-1} \}$.
$\mathbb{T}$ is a priority queue
sorted by the cost values as $\mathcal{J} = \lambda_t \cdot \mathcal{H}_t + \lambda_s \cdot \mathcal{H}_s + \lambda_e \cdot \mathcal{I}_e$,
where, $\mathcal{H}_t$ is the flight duration cost, $\mathcal{H}_s$ is the heuristic terminal cost, and $\mathcal{I}_e$ is the control efforts (detailed in Sec.\ref{sec: 3.4}).
The optimal trajectory ${\tau}_0$  is selected as the main branch for current navigation, while the others
serve as sub-branches, providing a set of alternative safe trajectories for emergencies.

\begin{figure}[!t]
	\centering
	\includegraphics[width=3.4in, trim=1 1 1 1, clip]{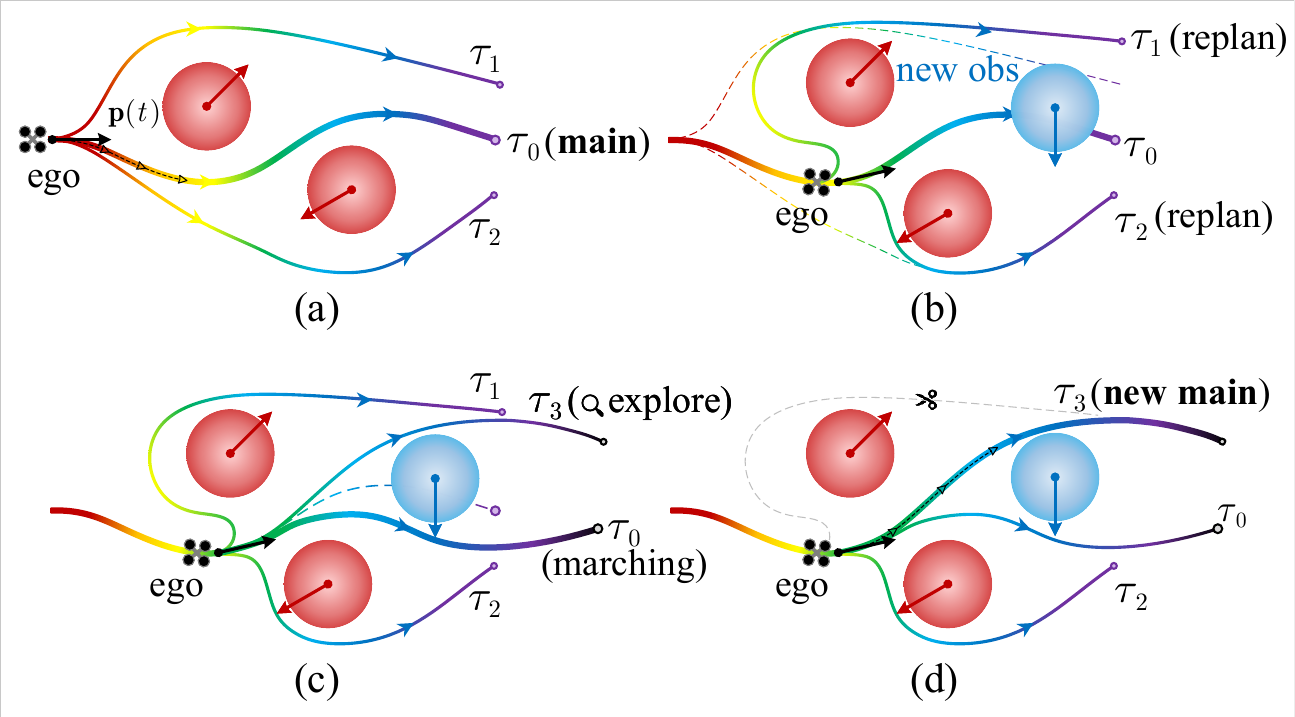}
	\caption{Incremental update process of topological trajectories.
		(a) The main $\tau_0$ publishes the navigation setpoints.
		(b) Replan of sub-branches $\tau_1$, $\tau_2$.
		(c) Forward planning of $\tau_0$, and exploration of new branches $\tau_3$.
		(d) Branches Decision. $\tau_3$ becomes the new main, while $\tau_1$ is cut off.
	}
	\vspace{-0.5cm}
	\label{fig: topological update}
\end{figure}

With the ego heading towards the goal, TRUST-Planner continuously monitors dynamic environmental changes and incrementally updates trajectory branches.
As shown in Fig.\ref{fig: topological update}, we employ parallel optimizers to simultaneously update and reconstruct trajectories in real-time:
1) {\it Forward marching of main branch}, which extends the terminals of main trajectory towards the goal; 2) {\it Backward replanning of sub-branches}, which updates the start of sub-branches with current states; and 3) {\it Forward exploration of new branches}, which explores
new spatial-temporal topologies for future navigation.
These processes fully utilize the historical
information to efficiently maintain the topological diversity of trajectory queue.

\subsection{Topological Path Planning} \label{sec: 3.3}

The topological path planning aims to explore the collision-free space in dynamic environments, and find diverse homotopy distinct reference paths for trajectory guidance.
Inspired by \cite{zhou_robust_2020}, we propose a Dynamic Enhanced Visible PRM ({\bf{DEV-PRM}}), which includes three key enhancements as follows.

\subsubsection{Predictive Directional Cone (PDC)}

\begin{figure}[!t]
	\centering
	\vspace{-0.5em}
	\includegraphics[width=3.49in]{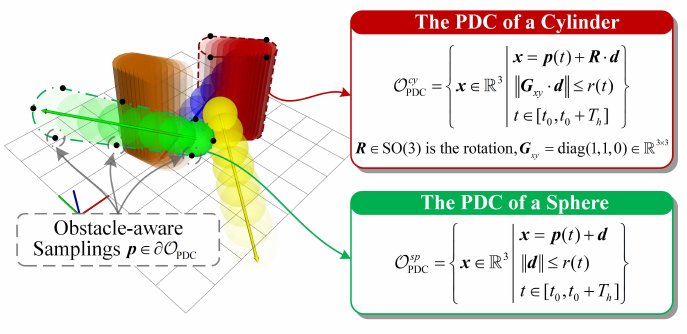}
	\caption{Predictive directional cones (PDCs) and obstacle-aware samplings.}
	\label{fig: PDC}
	\vspace{-0.5em}
\end{figure}

Traditional path planning methods lack explicit temporal information during the planning phase, which brings spatial-temporal inconsistency.
To address this issue, we project the predicted obstacle motion
into the spatial domain to incorporate temporal information in the path planning.
A geometric approximation, i.e., the Predictive Directional Cone (PDC), is constructed over a short time horizon to enable
dynamic collision checking.
As shown in Fig.\ref{fig: PDC}, given a dynamic obstacle $\mathcal{O}$, with intrinsic geometric size $r_o$, observed position ${\bf p}_o \in \mathbb{R}^3 $, and velocity ${\bf v}_o \in \mathbb{R}^3 $ at current time $t_0$.
Its future motion is ${\bf p}_o(t)={\bf p}_o+{\bf v}_o\cdot (t-t_0) + \delta(t)$, where $\delta(t)$ is the uncertainty maneuver term.
Over a short horizon $T_h$, assume that $\| \delta(t) \| \leq \mu_o \cdot (t-t_0)$ is linear bounded, where $\mu_o$ is the slope.
Then, the obstacle motion can be projected as the expansion of its geometric size $r_o(t) = r_o + \mu_o \cdot (t-t_0)$.
Therefore, the PDC can be defined as the union of cross-sectional volumes along ${\bf v}_0$ as
\begin{equation}
	\mathcal{O}_{\text{PDC}} = \bigcup \mathcal{O}[
		{\bf p}_o(t), r_o(t)],
	t \in [t_0, t_0 + T_h]
	\label{eqn: PDC}
\end{equation}
where $\mathcal{O}_{\text{PDC}}({\bf p},r)$ denotes the transformation of the base shape $\mathcal{O}$ with a translation and a geometric scaling by $\frac{r}{r_o}$. $\mathcal{O}_{\text{PDC}}({\bf p},r)$ is a conservative spatial estimate of obstacle occupancy over a short horizon, which implicitly encodes the time information to support spatial-temporal topological path exploration.

\subsubsection{Obstacle-Aware Sampling Strategy}

Since the approximate motions of dynamic obstacles have been encoded in PDCs, we can precompute a set of obstacle-aware samples to guide the initialization of the topological roadmap as $\mathcal{P}_{\text{PDC}} = \{ {\bf p} \in \partial \mathcal{O}_{\text{PDC}}\}$, where $\partial \mathcal{O}_{\text{PDC}}$ denotes the surface of $\mathcal{O}_{\text{PDC}}$.
These samples are selected on multiple slices along the predicted velocity ${\bf v}_0$.
The obstacle-aware samples $\mathcal{P}_{\text{PDC}}$ are inserted before the random sampling phase to initialize DEV-PRM, enhancing the roadmap's ability to capture spatial-temporal topological structures.

\begin{figure}[!t]
	\centering
	\vspace{-0.5em}
	\includegraphics[width=3.1in, trim=1 1 1 1, clip]{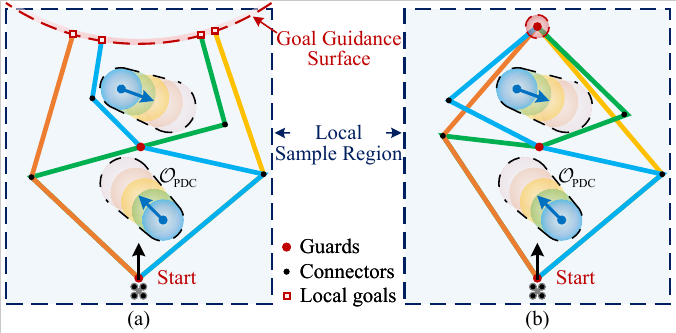}
	\caption{Illustration of Goal Guidance Surface (GCS). (a) The topological paths terminate at relaxed local goals on the GCS to improve optimality. (b) Near the global goal, the GCS degenerates to a single point.}
	\label{fig: GGC}
	\vspace{-0.5em}
\end{figure}

\begin{algorithm}[!t]
	\SetNlSty{}{}{}
	\SetAlgoNlRelativeSize{0}
	\SetAlgoLined
	\SetFuncSty{textbf}    
	\SetFuncArgSty{text}     
	\caption{DEV-PRM Algorithm} \label{alg: DEV-PRM}
	\SetKwProg{Fn}{Function}{:}{}
	\SetKwFunction{AddGuard}{AddGuard}
	\SetKwFunction{ConstructRoadMap}{ConstructRoadMap}
	\SetKwFunction{ConstructRoadMapRandomly}{ConstructRoadMapRandomly}
	\SetKwFunction{BeamSearch}{BeamSearch}
	\SetKwFunction{TopoPathSelect}{TopoPathSelect}
	\KwIn{${\bf p}_{s}$, ${\bf p}_{g}$, $\{\mathcal{O}_{\text{PDC}}\}$}
	$r_{\text{des}} \leftarrow \| {\bf p}_{s} - {\bf p}_{g}\| - T_h v_{\text{max}} $ \;
	Create the $\mathcal{S}_{\text{GGS}}$(${\bf p}_{g}$, $r_{\text{des}}$) \;
	$\mathcal{G}$.\AddGuard{${\bf p}_{s}$}, $\mathcal{G}$.\AddGuard{$\mathcal{S}_{\text{GGS}}$}\;
	\lFor{\rm each $\mathcal{O}_{\text{PDC}}$}{$\mathcal{P}_{\text{PDC}} \leftarrow  \mathcal{P}_{\text{PDC}} \cap \{{\bf p} \in \partial \mathcal{O}_{\text{PDC}}\}$}
	\ConstructRoadMap{$\mathcal{G}$, $\mathcal{P}_{\text{PDC}}$}\;
	\ConstructRoadMapRandomly{$\mathcal{G}$, $t_{\text{max}}$, $N_{\text{max}}$}\;
	${\bf \Pi}_{\text{raw}} \leftarrow$ \BeamSearch{$\mathcal{G}$}, ${\bf \Pi} \leftarrow$ \TopoPathSelect{${\bf \Pi}_{\text{raw}}$}
	\textbf{return} TopoPathSet ${\bf \Pi}$ \;
\end{algorithm}

\subsubsection{Goal Guidance Surface (GGS)}

In the original V-PRM \cite{zhou_robust_2020}, the topological paths tend to directly connect with a fixed local goal, resulting in lateral zigzagging motions, thereby degrading the optimality of the derived trajectories.
To address this issue, the Goal Guidance Surface (GGS) is introduced.
As shown in Fig.\ref{fig: GGC}, GGS replaces the local goal with a heuristic reference surface
$\mathcal{S}_{\text{GGS}} = \{ \|{\bf p} - {\bf p}_{g}\|_2 = r_{\text{des}} \}$,
where ${\bf p}_{\text{g}}$ is the global goal and $r_{\text{des}}$ is the cut-off radius.
$\mathcal{S}_{\text{GGS}}$ defines a spherical surface that stays a constant distance from the global goal.
During the sampling process, it acts as a special guard node, guiding paths towards the global goal.
It gradually shrinks into a single point as the drone approaches the global goal, naturally degenerating into the V-PRM.

Based on these enhancements, the proposed DEV-PRM is outlined in Algorithm \ref{alg: DEV-PRM}.
Given the current start ${\bf p}_{s}$, the global goal ${\bf p}_{\text{g}}$, and $\{\mathcal{O}_{\text{PDC}}\}$ of dynamic obstacles, it outputs the topological path set ${\bf \Pi}$,
which introduce diverse spatial-temporal homotopies for guiding subsequent trajectory optimization.

\subsection{Spatial-temporal Trajectory Optimization} \label{sec: 3.4}

For efficient parallel trajectory computation, we adopt a Uniform Terminal-Free MINCO ({\bf{UTF-MINCO}}) to parameterize the trajectory with piece-wise continuous polynomials
\begin{equation}
	\mathbf{p}(t) = \mathbf{C}_{i}^{\top} \cdot \mathbf{b}(t-t_{i-1}), t \in [t_{i-1}, t_{i}], i = 1,\ldots,n
	\label{eqn: Polynomials}
\end{equation}
where $\mathbf{b}(t) = [1, t, t^2, \ldots, t^m]^{\top} \in \mathbb{R}^{m+1}$ is the polynomial basis;
$\mathbf{C}_{i} \in \mathbb{R}^{(m+1) \times 3}$ is the coefficient.
$t_{i}$ is the intermediate time stamp for two adjacent segments, with the corresponding joint $\mathbf{p}(t_i) = \mathbf{p}_i$.
In particular, the initial and final times are denoted as $t_0$ and $t_f$, respectively, whose joints are $\mathbf{p}_0$ and $\mathbf{p}_f$.
To ensure the uniform distribution of control points, the time duration of each segment is set as $T_i = t_{i}-t_{i-1}=(t_f-t_0)/n$.
Denote $\mathbfcal{S}=[\mathbf{p}_f, \dot{\mathbf{p}}_f] \in \mathbb{R}^{3 \times 2}$ as the free terminal states, $\mathbfcal{P} = [\mathbf{p}_1, \mathbf{p}_2, \ldots, \mathbf{p}_i,\ldots, \mathbf{p}_{n-1}] \in \mathbb{R}^{3 \times (n-1)}$ as the intermediate control points, and $T_{\Sigma} = \sum_{i = 1}^{n} T_i$ as the total time. 
The UTF-MINCO formulates \eqref{eqn: Polynomials} into a linear-time and space complex mapping form as $\mathbfcal{C} = \mathbfcal{A}^{-1}(T_{\Sigma}) \cdot \mathbfcal{D}(\mathbfcal{S}, \mathbfcal{P})$,
where $\mathbfcal{C}={\rm col}\{\mathbf{C}_1, \ldots, \mathbf{C}_{n}\} \in \mathbb{R}^{6n \times 3}$;
$\mathbfcal{A}$ is a non-singular band matrix; and $\mathbfcal{D}$ is the boundary matrix (see \cite{wang_geometrically_2022}).


Then, the trajectory planning is formulated as the following unconstrained optimization
\begin{equation}
	\underset{\scriptscriptstyle \mathbfcal{S}, \mathbfcal{P}, T_{\Sigma}}{\textbf {min}} \quad
	\mathcal{J} = \mathcal{H}(\mathbfcal{S}, T_{\Sigma}) + \mathcal{I}[\mathbfcal{C}(\mathbfcal{S}, \mathbfcal{P}, T_{\Sigma}), T_{\Sigma}]
	\label{eqn: unconstrained optimization}
\end{equation}
in which, $\mathcal{H}$ denotes the cost term that directly acts on the $\mathbfcal{S}$ and $T_{\Sigma}$, while $\mathcal{I}$ represents the integral cost defined among the polynomials. The analytical gradients of \eqref{eqn: unconstrained optimization} can be derived to facilitate efficient iteration in the optimization process, as
\begin{equation}
	\frac{\partial \mathcal{J}}{\partial {\mathbfcal{S}}} = \frac{\partial \mathcal{H}}{\partial {\mathbfcal{S}}} + \frac{\partial \mathcal{I}}{\partial {\mathbfcal{S}}},
	\frac{\partial \mathcal{J}}{\partial {\mathbfcal{P}}} = \frac{\partial \mathcal{I}}{\partial {\mathbfcal{P}}},
	\frac{\partial \mathcal{J}}{\partial {T_{\Sigma}}} = \frac{\partial \mathcal{H}}{\partial T_{\Sigma}} + \frac{1}{n} \sum_{i = 1}^{n}\frac{\partial \mathcal{I}}{\partial {T_i}}
	\label{eqn: analytical gradients}
\end{equation}
where $\frac{\partial \mathcal{I}}{\partial {\mathbfcal{P}}}$ and $\frac{\partial \mathcal{I}}{\partial {T_i}}$ follow the spatial-temporal gradient conversion of the original MINCO \cite{wang_geometrically_2022}. Differently, in \eqref{eqn: analytical gradients}, the term $\frac{1}{n} \sum_{i = 1}^{n}\frac{\partial \mathcal{I}}{\partial {T_i}}$  arises due to the uniform discretization of the total time $T_{\Sigma}$.
On the other hand, $\frac{\partial \mathcal{I}}{\partial \mathbfcal{S}}$, i.e., the gradient with respect to the terminal states $\mathbfcal{S}$, is derived as
\begin{equation}
	\frac{\partial \mathcal{I}}{\partial \mathbfcal{S}} =
	\begin{bmatrix} \mathbfcal{G}_{[6n-2, :]} \\ \mathbfcal{G}_{[6n-1, :]}
	\end{bmatrix}^{\top}, \quad
	\mathbfcal{G} = \mathbfcal{A}^{-\top} \cdot \frac{\partial \mathcal{I}}{\partial {\mathbfcal C}} \in \mathbb{R}^{6n \times 3}
	\label{eqn: analytical gradients I to S}
\end{equation}
where the notation $\mathbfcal{G}_{[i, :]}$ denotes the $i^\text{th}$ row of $\mathbfcal{G}$.

In this paper, the objective $\mathcal{J}$ in \eqref{eqn: unconstrained optimization} is designed as
\begin{equation}
	\mathcal{J} = [\mathcal{H}_t, \mathcal{H}_s,
	\mathcal{I}_{o}, \mathcal{I}_d, \mathcal{I}_e] \cdot {\boldsymbol{\lambda}}
	\label{eqn: design of J}
\end{equation}
in which, ${\boldsymbol{\lambda}} \in \mathbb{R}^5$ is the weight vector, and the corresponding cost terms $\mathcal{H}_t$, $\mathcal{H}_s$,
$\mathcal{I}_{o}$, $\mathcal{I}_d$, $\mathcal{I}_e$ are detailed as follows.

\subsubsection{Flight duration cost \texorpdfstring{$\mathcal{H}_t$}{Ht}}
The cost $\mathcal{H}_t = T_{\Sigma}$ is introduced to minimize the flight time for fast traversal. The gradients are given by $\frac{\partial \mathcal{H}_t}{\partial T_{\Sigma}} = 1$.

\subsubsection{Heuristic terminal guidance cost \texorpdfstring{$\mathcal{H}_s$}{Hs}}
To guide the drone to approach the given global goal position $ \mathbf{p}_g \in \mathbb{R}^3$, the cost $\mathcal{H}_s$ is designed as the following heuristical form
\begin{equation}
	\begin{split}
		 & \mathcal{H}_s                                      =
		\|\mathbf{p}_f - \mathbf{p}_{g}\|_{{\bf K}_f^p} +
		\alpha \|\dot{\mathbf{p}}_f\|_{{\bf K}_f^{\dot p}}           \\
		 & \frac{\partial \mathcal{H}_s}{\partial {\mathbf{p}}_f}  =
		\frac{{\bf K}_f^p \cdot (\mathbf{p}_f - \mathbf{p}_{g})}{\|\mathbf{p}_f - \mathbf{p}_{g}\|_{{\bf K}_f^p}}, \quad
		\frac{\partial \mathcal{H}_s}{\partial \dot{\mathbf{p}}_f}  =
		\frac{\alpha {\bf K}_f^{\dot p} \cdot \dot{\mathbf{p}}_f}{\|\dot{\mathbf{p}}_f\|_{{\bf K}_f^{\dot p}}}
	\end{split}
	\label{eqn: design of Hs}
\end{equation}
where, $\|\mathbf{x}\|_{\mathbf{K}} = \sqrt{\mathbf{x}^{\top}\mathbf{K}\mathbf{x}}$;
$\mathbf{K}_f^p, \mathbf{K}_f^{\dot p} \in \mathbb{R}^{3\times3}_{+}$ are the weighting matrices for the terminal errors. $\alpha$ serves as an adjustment coefficient, which is set to 0 when the drone is far from the global goal. As the drone approaches, let $\alpha = 1-{\| {\bf p}_0 - {\bf p}_g \|}/{d_{\varepsilon}}$ to reduce the terminal velocity error, where $d_{\varepsilon}$ is a threshold.

As for the integral objective $\mathcal{I}$, we utilize the trapezoidal method to approximate its value 
as the sum of the integrand $\mathcal{L}(\mathbf{C}_i, t)$ evaluated at discrete sample time $t_k$, given by
\begin{equation}
	\mathcal{I}[\mathcal{L}] = \sum_{i = 1}^{n} \mathcal{I}_i
	= \sum_{i = 1}^{n} \sum_{k = 0}^{\kappa} \omega_k \cdot \frac{T_i}{\kappa} \cdot \mathcal{L}(\mathbf{C}_i, t_k)
	\label{eqn: trapezoidal integration}
\end{equation}
in which, $\mathcal{I}_i$ is the objective value for the $i^\text{th}$ segment.
$\kappa$ is the number of sampling intervals.
$t_k = k\frac{T_i}{\kappa} + \sum_{j = 1}^{i-1} T_j$ is the discrete sample time stamp.
$\omega_k$ denotes the trapezoidal coefficient, taking the value as $\frac{1}{2}$ for $k=0$ or $k=\kappa$, and $1$ otherwise.
The partial derivatives of $\mathcal{I}_i$ are given by
\begin{subequations}
	\label{eqn: analytical gradients of I}
	\begin{align}
		 & \frac{\partial \mathcal{I}_i}{\partial \mathbf{C}_i} =
		\sum_{k = 0}^{\kappa} \omega_k \cdot \frac{T_i}{\kappa} \cdot \frac{\partial \mathcal{L}}{\partial \mathbf{C}_i}    \label{eqn: dI-a}                                                                \\
		 & \frac{\partial \mathcal{I}_i}{\partial T_i} =
		\frac{\mathcal{I}_i}{T_i} + \sum_{k = 0}^{\kappa} \omega_k \cdot \frac{T_i}{\kappa} \cdot \frac{k}{\kappa}\cdot \frac{\partial \mathcal{L}}{\partial t} \label{eqn: dI-b}                            \\
		 & \frac{\partial \mathcal{I}_i}{\partial T_j} = \sum_{k = 0}^{\kappa} \omega_k \cdot \frac{T_i}{\kappa} \cdot \frac{\partial \mathcal{L}}{\partial t}, \; j = 1, 2, \ldots, i - 1 \label{eqn: dI-c}
	\end{align}
\end{subequations}
Notice that, the term \eqref{eqn: dI-c} does not appear in the original MINCO \cite{wang_geometrically_2022}. It arises due to the time coupling among polynomial segments, e.g., for dynamic obstacle avoidance  $\mathcal{I}_o$, the time alignment between moving obstacles and the ego must be considered, introducing the additional coupling term \eqref{eqn: dI-c}.

The integral costs $\mathcal{I}_{o}$, $\mathcal{I}_d$ and $\mathcal{I}_e$ are detailed as follows.

\subsubsection{Obstacle avoidance penalty \texorpdfstring{$\mathcal{I}_o$}{Io}}

\begin{figure}[!t]
	\centering
	\vspace{-1.0em}
	\includegraphics[width=2.8in, trim=1 1 1 1, clip]{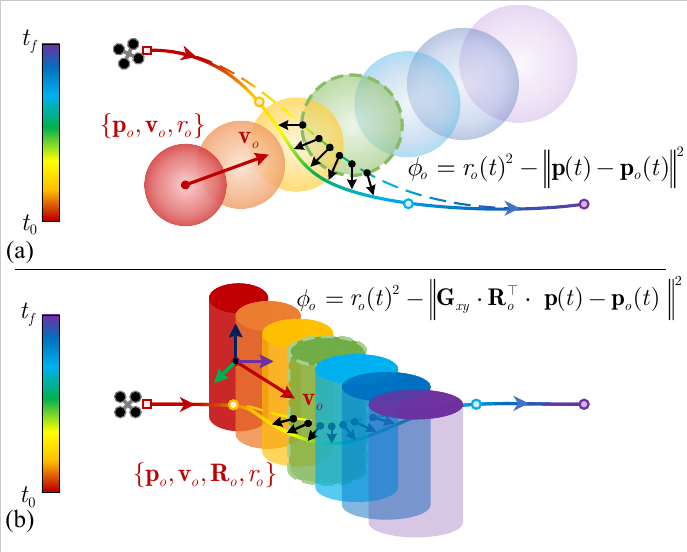} 
	\vspace{-0.5em}
	\caption{Illustrations of distance field (DDF) for obstacle avoidance.
		(a) DDF for a sphere obstacle.
		(b) DDF for a cylinder obstacle.
	}
	\vspace{-1.5em}
	\label{fig: DDF}
\end{figure}

To achieve both spatial and temporal obstacle avoidance, as shown in Fig.\ref{fig: DDF}, we introduce the {\bf D}ynamic {\bf D}istance {\bf F}ield ({\bf{DDF}}).
Based on the observed position $\mathbf{p}_o$ and velocity ${\bf v}_o$ at time $t_0$, the motion of a dynamic obstacle
is predicted as ${\mathbf{p}_o}(t)$, while uncertainty of its motion is reflected by a time-varying increasing avoidance radius $r_o(t) = r_o + r_{\rm safe} + \mu_o \cdot (t-t_0)$, where $r_o$ is the intrinsic size of the obstacle, $r_{\rm safe}$ is the additional safety margin, and $\mu_o$ denotes the uncertainty provided by the dynamic obstacle estimator.
Let $\phi_o$ as the signed squared distance between the ego trajectory with respect to the predicted obstacle surface.
$\phi_o$ becomes positive when the trajectory violates the inflated avoidance region.
Thus, based on the DDF, the following penalty is designed to ensure spatial-temporal avoidance as
\begin{equation}
	\mathcal{L}_o = \sum_{\mathcal{O}} f_{\text{max}} \{ \phi_o [ \mathbf{p}(t), \mathbf{p}_o(t), r_o(t)] \}
	\label{eqn: design of Io}
	\vspace{-0.7em}
\end{equation}
where $f_{\text{max}}(x)$ is a smooth approximation of $\text{max}(x, 0)$ as described in \cite{wang_robust_2022}.
The derivatives of $\phi_o$ are given as follows
\begin{equation}
	\vspace{-0.8em}
	\begin{split}
		\frac{\partial \phi_o}{\partial \mathbf{C}_i} & =  \mathbf{b} \cdot \left(\frac{\partial \phi_{o}}{\partial {\mathbf{p}}}\right)^{\top}                                                    \\
		\frac{\partial \phi_o}{\partial t}            & = \underbrace{\left(\frac{\partial \phi_o}{\partial {\mathbf{p}}}\right)^{\top} \cdot \dot{{\mathbf{p}}}}_{\text{(a) derivative of ego}} +
		\underbrace{\left(\frac{\partial \phi_o}{\partial {\mathbf{p}}_o}\right)^{\top} \cdot \dot{{\mathbf{p}}}_o +
			\frac{\partial \phi_o}{\partial r_o} \cdot \dot{r}_o}_{\text{(b) derivative of dynamic obstacles}}
	\end{split}
	\label{eqn: gradients of Io}
\end{equation}
Note that, $\frac{\partial \phi_o}{\partial t}$ contains two terms, the derivative of the ego's trajectory (\ref{eqn: gradients of Io}a), and the additional derivative of the dynamic obstacle (\ref{eqn: gradients of Io}b).
As described in \eqref{eqn: analytical gradients of I}, (\ref{eqn: gradients of Io}a) only affects $\frac{\partial \mathcal{I}_i}{\partial T_i}$ of $i^\text{th}$ trajectory segment.
In contrast, the remaining (\ref{eqn: gradients of Io}b) arises from the alignment between the ego trajectory and the dynamic obstacle, thereby introducing additional temporal coupling that contributes to both ${\partial \mathcal{I}_i}/{\partial T_i}$ and ${\partial \mathcal{I}_i}/{\partial T_j}$ for $j = 1, \ldots, i-1$.

\subsubsection{Dynamic feasibility cost \texorpdfstring{$\mathcal{I}_d$}{I d}}
The speed and acceleration are limited by the physical capabilities.
Hence, $\mathcal{I}_d$ is designed by penalizing $\mathcal{L}_d^v = f_{\max}(\phi_v)$ and $\mathcal{L}_d^a = f_{\max}(\phi_a)$ as
\begin{equation}
	\phi_v = {\mathbf{v}^{\top}\mathbf{v}}/{v_{\max}^2}-1, \quad
	\phi_a = {\mathbf{a}^{\top}\mathbf{a}}/{a_{\max}^2}-1
	\label{eqn: design of Id}
\end{equation}
in which, $v_{\max}$ and $a_{\max}$ denote the maximum speed and acceleration, respectively.
The gradients of \eqref{eqn: design of Id} are given by
\begin{equation}
	\begin{split}
		{\partial \phi_v}/{\partial \mathbf{C}_i} & = {2 \dot{\mathbf{b}} \mathbf{v}^{\top}}/{v_{\max}^2}, \quad
		{\partial \phi_v}/{\partial t} = {2 \mathbf{a}^{\top} \mathbf{v}}/{v_{\max}^2}                            \\
		{\partial \phi_a}/{\partial \mathbf{C}_i} & = {2 \ddot{\mathbf{b}} \mathbf{a}^{\top}}/{a_{\max}^2}, \quad
		{\partial \phi_a}/{\partial t} = {2 \mathbf{j}^{\top} \mathbf{a}}/{a_{\max}^2}
	\end{split}
	\label{eqn: gradients of Id}
\end{equation}
where $\mathbf{j} = \dddot{\mathbf{p}}  \in \mathbb{R}^3$ denotes the jerk vector.

\subsubsection{Jerk minimization cost \texorpdfstring{$\mathcal{I}_e$}{I e}}
$\mathcal{I}_e$ penalizes the jerk to promote smoother motions and reduce the control efforts. The corresponding integrand $\mathcal{L}_e$ and its gradients are given by
\begin{equation}
	\mathcal{L}_e = \mathbf{j}^{\top}\mathbf{j}, \quad
	\frac{\partial \mathcal{L}_e}{\partial \mathbf{C}_i} = 2\dddot{\mathbf{b}}^{\top}\mathbf{j}, \quad
	\frac{\partial \mathcal{L}_e}{\partial t} = 2\mathbf{s}^{\top}\mathbf{j}
	\label{eqn: design of Ie}
\end{equation}
where $\mathbf{s} = \ddddot{\mathbf{p}} \in \mathbb{R}^3$ denotes the snap vector.

In summary, with the above problem formulation, the optimization is transcribed as a lightweight unconstrained nonlinear programming as shown in \eqref{eqn: unconstrained optimization}.
The flight trajectories are compactly parameterized by lower-dimensional
$\mathbfcal{S}$, $\mathbfcal{P}$ and $T_{\Sigma}$.
Owing to the analytical gradients, it requires only a single function evaluation per gradient iteration, resulting in linear time and space complexity.
Therefore, using gradient-based solvers such as L-BFGS \cite{liu_limited_1989a}, \eqref{eqn: unconstrained optimization} enables efficient parallel computation of multiple trajectories with different spatial-temporal topologies to explore diverse solutions in real time.

\section{Simulation Experiments and Results}

This section presents several simulation experiments to validate the performance of the proposed method.
The software of TRUST-Planner is implemented by C++ under ROS 2 (Humble) \cite{steven_robot_2022} environment on a desktop computer with a 2.2 GHz Intel Core i9-13980HX CPU and 16 GB RAM.
The simulation drone follows the planned trajectories by a nonlinear controller \cite{mellinger_minimum_2011} with constraints $v_{\max} = 10 \, \mathrm{m/s}$ and $a_{\max} = 15 \, \mathrm{m/s^2}$.
Note that, the planner can not access to the precise future motion of dynamic obstacles, but only the current position and velocity to estimate by utilizing the Extended Kalman Filter (EKF) \cite{wan_ekf_2000}.
The details are as follows.

\vspace{-1em}
\subsection{Topological Path Planning} \label{sec: 4.1}

\begin{figure}[!h]
	\centering
	\vspace{-1em}
	\includegraphics[width=3.35in]{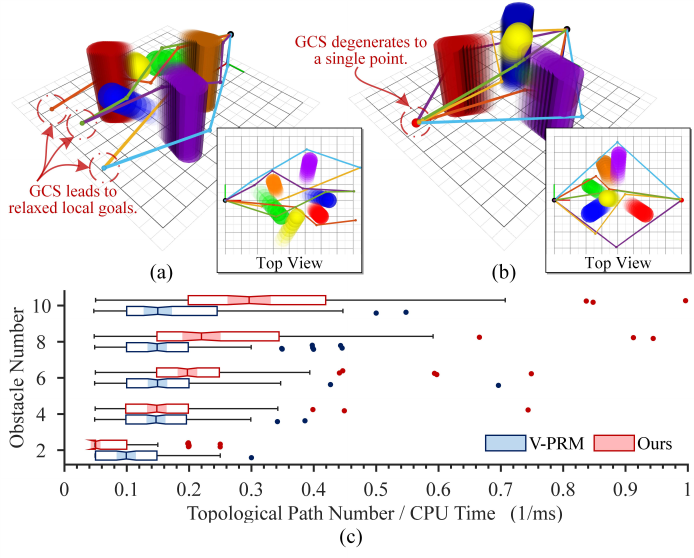} 
	\vspace{-1em}
	\caption{Topological path planning results of DEV-PRM. (a) $\mathcal{S}_{\text{GGS}}$ with ${\bf p}_{g} = [65, 0, 7.5] ^\top \, \mathrm{m}$ and $r_{\text{des}} = 25 \, \mathrm{m}$. (b) $\mathcal{S}_{\text{GGS}}$ with ${\bf p}_{g} = [40, 0, 7.5] ^\top \,  \mathrm{m}$ and $r_{\text{des}} = 0 \, \mathrm{m}$. (c) Comparison of the capabilities to find distinct topological paths (found path number / CPU Time).}
	\label{fig: Sim-A-1-1}
\end{figure}

We first compare the proposed DEV-PRM with the original V-PRM \cite{zhou_robust_2020}\footnote{V-PRM is reimplemented on C++ under ROS 2 environment.} to demonstrate its effectiveness for topological path planning in dynamic environments.
The test scenarios are set within a $20\times20\times10\,\mathrm{m^3}$ area containing randomly generated dynamic obstacles.
Both methods are limited to a maximum computation time of $t_{\rm max} = 20\,\mathrm{ms}$.
As shown in Fig.\ref{fig: Sim-A-1-1}, DEV-PRM successfully generates homotopy-distinct paths that avoid dynamic obstacles.
When the GGS is set far from the goal ${\bf p}_{g}$, DEV-PRM produces topological paths with relaxed local endpoints. As the GGS is close to the goal, it shrinks to a single point, causing DEV-PRM to degenerate into the original behavior of V-PRM with fixed local endpoints.
Fig.\ref{fig: Sim-A-1-1}(c) further provides a quantitative comparison of the efficiency (found homotopy-distinct path number/CPU time) under varying numbers of random obstacles.
Across all the tests, benefiting from the obstacle-aware sampling strategy, DEV-PRM consistently discovers more topological paths than V-PRM in limited CPU times.
Specifically, with 10 random dynamic obstacles, DEV-PRM achieves an average of 0.338, compared to 0.186 for V-PRM, representing an 81.9\% improvement.
The results validate the superior efficiency of the topological diversity of DEV-PRM in dynamic environments.

\subsection{Trajectory Optimization} \label{sec: 4.2}

\begin{figure}[!t]
	\centering
	\includegraphics[width=3.40in]{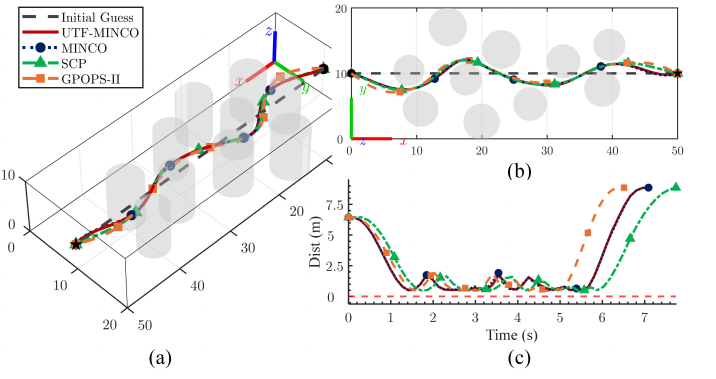}
	\vspace{-1em}
	\caption{Companions of trajectory optimization methods in a static scenario.
		(a) Trajectories.
		(b) Top view. 
		(c) Distance to the obstacles.}
	\label{fig: Sim-A-2-1}
	\vspace{-1em}
\end{figure}

\begin{figure}[!t]
	\centering
	\includegraphics[width=3.40in]{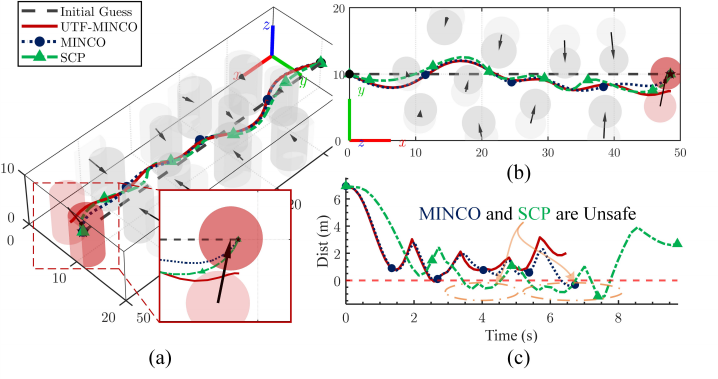}
	\vspace{-1em}
	\caption{Companions of trajectory optimization methods in a dynamic scenario.
		(a) Trajectories.
		(b) Top view. 
		(c) Distance to the obstacles.}
	\label{fig: Sim-A-2-2}
	\vspace{-1em}
\end{figure}

\begin{table}[!t]
	\renewcommand\arraystretch{1.2}
	\begin{center}
		\caption{Comparisons of Different Trajectory Optimization Methods}
		\vspace{-1em}
		\label{tab: Sim-A-2}
		\begin{tabular}{ccclcc}
			\toprule
			\multirow{2}{*}{\bf Method}            & \multicolumn{2}{c}{\bf Static CASE}              &                                                 & \multicolumn{2}{c}{\bf Dynamic CASE}                                                                                                      \\ \cline{2-3} \cline{5-6}
			                                       & {\fontsize{6pt}{7.2pt}\selectfont CPU Time (ms)} & {\fontsize{6pt}{7.2pt}\selectfont Duration (s)} &                                      & {\fontsize{6pt}{7.2pt}\selectfont CPU Time (ms)} & {\fontsize{6pt}{7.2pt}\selectfont Duration (s)} \\ \cline{1-6}
			UTF-MINCO                              & 8.98                                             & 7.08                                            &                                      & 9.71                                             & 6.47                                            \\
			MINCO \cite{wang_geometrically_2022}   & 7.77                                             & 7.09                                            &                                      & 8.52                                             & 6.72                                            \\
			SCP \cite{long_Realtime_2023}          & 301.7                                            & 7.40                                            &                                      & 511.4                                            & 9.72                                            \\
			GPOPS-II \cite{patterson_gpopsii_2014} & 8127.3                                           & 6.51                                            &                                      & Failed                                           & Failed                                          \\
			\bottomrule
		\end{tabular}
	\end{center}
	\vspace{-0.8cm}
\end{table}

The performance of the proposed UTF-MINCO is compared with several trajectory optimization methods, including the original MINCO\cite{wang_geometrically_2022}\footnote{MINCO is reimplemented on C++ under ROS 2 environment.},
SCP \cite{long_Realtime_2023} solved by ECOS \cite{domahidi_ECOS_2013}, and GPOPS-II \cite{patterson_gpopsii_2014}\footnote{SCP and GPOPS-II are implemented on C-MEX under MATLAB.}.
As illustrated in Fig.\ref{fig: Sim-A-2-1}--\ref{fig: Sim-A-2-2}, two $50\times20\times10\,\mathrm{m^3}$ scenarios are designed to evaluate these methods.
In the static scenario, all methods successfully generate feasible trajectories to avoid the obstacles.
However, in more challenging dynamic scenario, where the moving obstacles gradually block the trajectories and the terminal states (red obstacle), only ours successfully avoids all dynamic obstacles owing to the designed DDF and relaxed terminal formulation.
Notably, GPOPS-II failed to get a feasible trajectory within a reasonable CPU time.
The average CPU Time and trajectory duration are summarized in Table \ref{tab: Sim-A-2}.
In terms of optimality, these methods have similar flight durations.
However, ours achieves millisecond-level computation time, which is much faster than SCP and GPOPS-II.
Despite it taking slightly more time than MINCO due to the additional optimization variables introduced by the relaxed terminal,
this overhead brings the improvement of the dynamic obstacle avoidance flexibility.

Overall, the results demonstrate that our method provides greater flexibility, safety, and computational efficiency in dynamic environments, enabling subsequent parallel optimization of multiple topological trajectories in real-time.

\subsection{Spatial-temporal Topological Trajectory Planning} \label{sec: 4.3}

\begin{table}[!t]
	\renewcommand\arraystretch{1.2}
	\begin{center}
		\caption{Comparisons of Spatial-temporal Topological Trajectories}
		\vspace{-0.2cm}
		\label{tab: Sim-B-1}
		\begin{tabular}{c|cccc|c}
			\toprule
			{\bf Idx} & $\mathcal{H}_t$ & $\mathcal{H}_s$ & $\mathcal{I}_e$                        & $\mathcal{J}$ & CPU Time (ms) \\ \hline
			$\tau_0$  & {\bf 4.39}      & 25.26           & 1.29$\times \text{10}^{\text 3}$       & {\bf 7.04}    & 9.88          \\
			$\tau_1$  & 4.58            & \bf{25.24}      & 1.35$\times \text{10}^{\text 3}$       & 7.24          & 8.12          \\
			$\tau_2$  & 4.86            & 25.25           & {\bf 8.22$\times \text{10}^{\text 2}$} & 7.47          & {\bf 4.10}    \\
			$\tau_3$  & 4.87            & 25.27           & 2.07$\times \text{10}^{\text 3}$       & 7.60          & 7.14          \\
			$\tau_4$  & 5.65            & 25.25           & 2.73$\times \text{10}^{\text 3}$       & 8.45          & 5.43          \\
			\bottomrule
		\end{tabular}
	\end{center}
	\vspace{-0.65cm}
\end{table}

\begin{figure*}[!htp]
	\centering
	\includegraphics[width=0.95\textwidth]{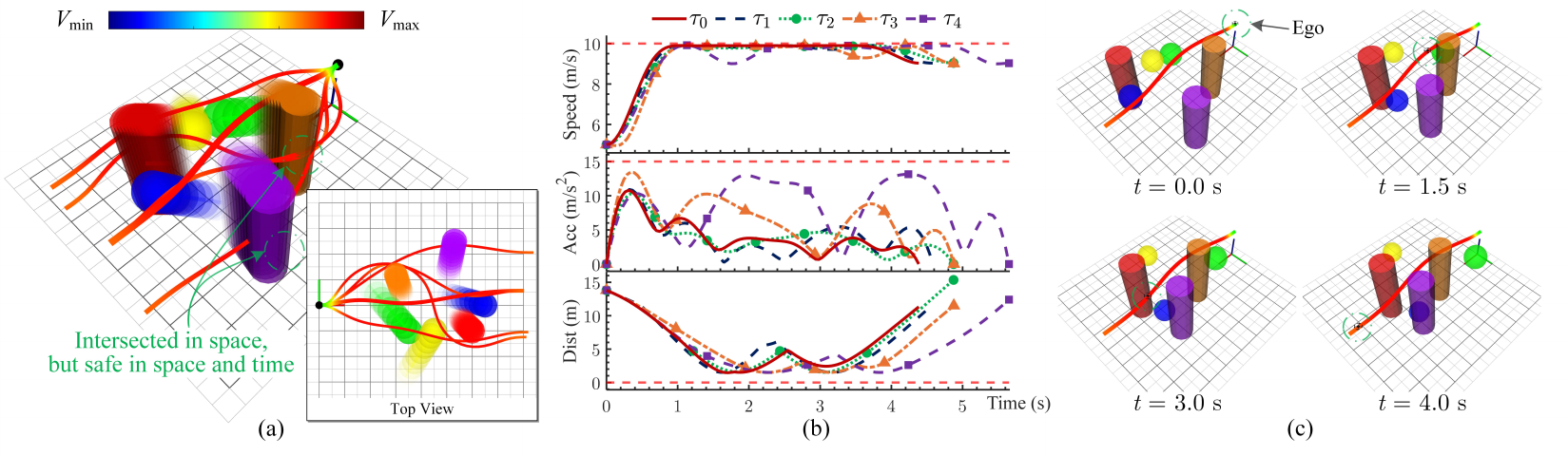} 
	\caption{Spatial-temporal topological trajectory planning results.
		(a) Spatial-temporal topological trajectories.
		(b) The speed profiles, acceleration and minimal distance to the obstacles.
		(c) Following process of the optimal trajectory $\tau_0$.}
	\label{fig: spatial-temporal topological planning}
\end{figure*}

This subsection analyzes the spatial-temporal topological trajectory planning results of TRUST-Planner.
The planning scenario is similar to Fig.\ref{fig: Sim-A-1-1}(a) in Sec.\ref{sec: 4.1}.
By leveraging the DEV-PRM as an initial guide, TRUST-Planner generates multiple distinct spatial-temporal topological trajectories.
The trajectories are sorted by $\mathcal{J}$, with $\lambda_t = 1$, $\lambda_s = 1 / v_{\rm max}$
and $\lambda_e=10^{-4}$ (refers to \cite{li_differential_2025a}).
The planning results are shown in Fig.\ref{fig: spatial-temporal topological planning}.
Five homotopy-distinct trajectories are generated, denoted as $\tau_0$--$\tau_4$.
Notice that, although some trajectories intersect with the PDCs of dynamic obstacles in space, they are all collision-free in the spatial-temporal domain.
The speed, acceleration, and minimal distance to the obstacles are all within the corresponding constraints.
The quantitative comparisons of these trajectories are provided in Table \ref{tab: Sim-B-1}.
Among them, the trajectory $\tau_0$ achieves the lowest weighted cost $\mathcal{J}$, and thus is selected as the optimal main branch for the ego drone to follow and execute.
Fig.\ref{fig: spatial-temporal topological planning}(c) illustrates the trajectory-following process, where the ego drone rapidly avoids dynamic obstacles and continues its forward flight.
Notice that, owing to the high efficiency of the low-level trajectory optimization, the proposed method can rapidly compute all these trajectories in parallel, with a maximum CPU time of 9.88 ms, fully satisfying the real-time requirements of practical applications.
Moreover, all generated trajectories are feasible and collision-free, providing a set of safe alternatives for emergency obstacle avoidance, which further enhances the robustness and flexibility of the proposed planner.

\subsection{Comparative Studies with Baseline Methods} \label{sec: 4.4}

In this subsection, the overall performance of TRUST-Planner is compared with several baseline methods.
We design challenging scenarios with dimensions of $250 \times 75 \times 20\,\mathrm{m}^3$, densely filled with static point clouds,
as well as 100 randomly generated dynamic obstacles.
All these baselines employ our UTF-MINCO
backend for dynamic obstacle avoidance.
The main distinction lies in the hierarchical guidance strategies as:

\begin{itemize}
	\item \textbf{Baseline 1}: No hierarchical guidance. Use a straight line towards the goal as the trajectory's initial guess.
	\item \textbf{Baseline 2}: Use the shortest path computed by Sparse A* \cite{long_Realtime_2023} as the initial guess for trajectory optimization. It can be regarded as a single topological guidance method.
	\item \textbf{Baseline 3}: Referring to \cite{zhou_robust_2020}, employ the DEV-PRM to generate multiple Topological paths for parallel trajectory optimization. However, it does not maintain historical topological information during replanning.
	\item \textbf{Ours}: The proposed TRUST-Planner integrates topological guidance with multi-branch management, enabling robust and efficient spatial-temporal trajectory planning.
\end{itemize}

\begin{figure*}[!htp]
	\centering
	\includegraphics[width=0.9\textwidth, trim=1 1 1 1, clip]{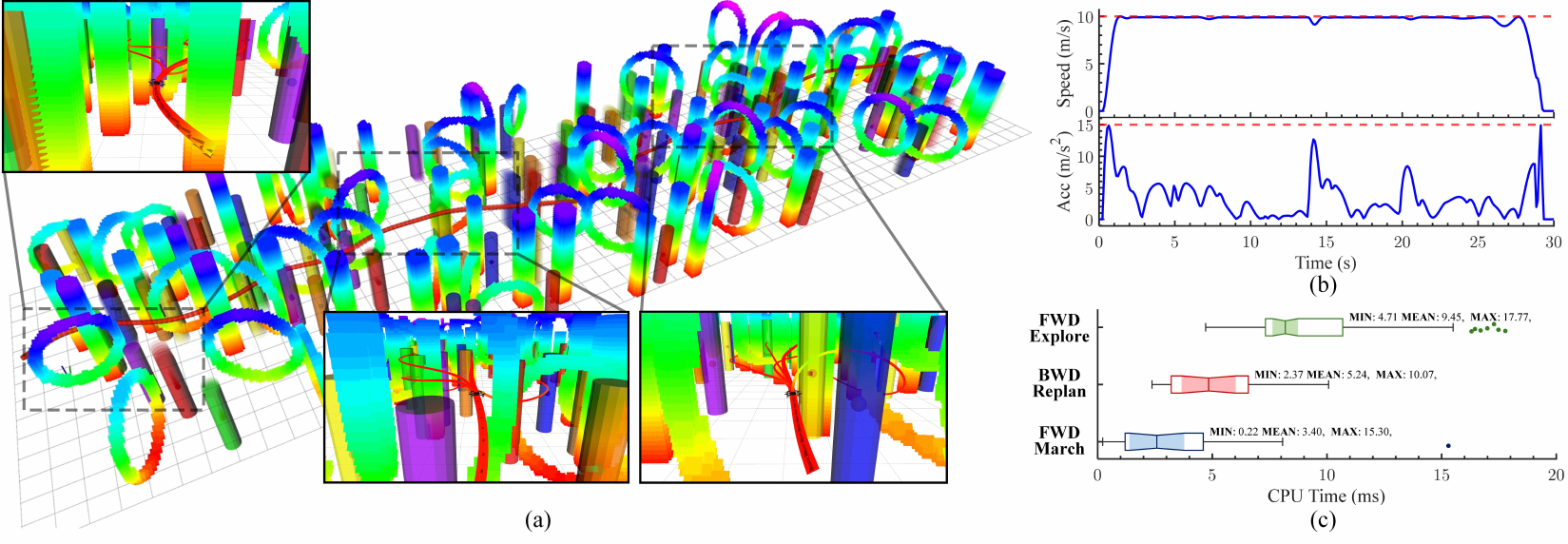} 
	\caption{Trajectory planning results of the proposed TRUST-Planner.
		(a) Flight processes, in which the enlarged sub-figures highlight the local multiple-branch topological trajectories.
		(b) The speed profiles and acceleration.
		(c) CPU time statistics of the incremental planning process.}
	\label{fig: baseline test 1}
\end{figure*}

\begin{figure}[!t]
	\centering
	\vspace{-0.5em}
	\includegraphics[width=3.49in, trim=1 1 1 1, clip]{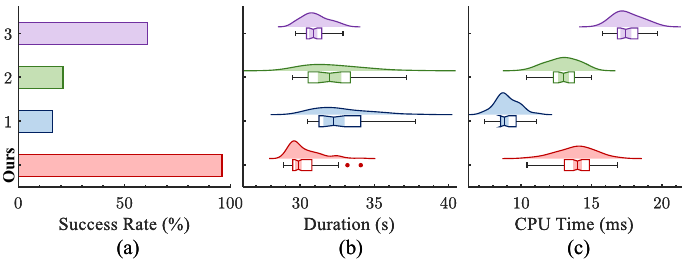} 
	\vspace{-1em}
	\caption{Statistical comparisons of competitors.
		(a) Success rate of safe arrival.
		(b) Arrival flight duration (optimality).
		(c) Average CPU time.}
	\label{fig: baseline test 3}
	\vspace{-1.5em}
\end{figure}

For each baseline, 100 times Monte-Carlo simulations are conducted to statistically evaluate the performance in terms of success rate, computation time, and arrival time (optimality).
The results are illustrated in Fig.\ref{fig: baseline test 1}--\ref{fig: baseline test 3}.
The detailed flight processes are provided in the supplementary video\footnote{[Online]. \href{https://www.bilibili.com/video/BV1RJWqzqEMz/}{https://www.bilibili.com/video/BV1RJWqzqEMz/}. Part 1.}.
Fig.\ref{fig: baseline test 1} shows the planning result. 
Our approach ensures safe and fast navigation through static and dynamic obstacles.
The ego drone maintains near-maximum speed to traverse the obstacle-rich area, quickly completing the flight in 29.4 s.
Fig.\ref{fig: baseline test 3} presents the statistical comparisons of these competitors.
Notice that, the planners can not access the precise future motion of dynamic obstacles but rely on estimations for collision avoidance.
In such challenging scenarios, Baseline 1, 2 are highly prone to becoming trapped in local minima and unable to escape, leading to high collision risks.
Baseline 3 employs a static multi-topology guidance strategy, enabling a little improvement of safety.
However, due to the lack of effective topological branch management, it may lose historical explored information during replanning, leading to suboptimal and eventual collision.
As a result, the baselines exhibit low success rates of safe arrival, while ours achieves a higher success rate of 96\%.
In terms of optimality, ours leverages optimal selection of spatial-temporal topological to maintain high-speed traversal in the obstacle area, achieving an average flight duration of 29.4~s, which is faster than all the competitors.
As for efficiency, since Baseline 2, 3 and ours require additional frontend path planning, they take longer CPU time for trajectory generation than Baseline 1.
However, owing to the incremental multi-branches topological management and parallel computation strategy, our method enables rapid trajectory generation at the 10~ms level, and reduces the average planning time of a single trajectory by 19.72\% compared to Baseline 3.
In Summary, the proposed TRUST-Planner demonstrates superior performance in terms of robustness, optimality, and computational efficiency.

\section{Real-world Experiments}

\begin{figure*}[!htp]
	\centering
	\vspace{-2.0em}
	\includegraphics[width=0.95\textwidth]{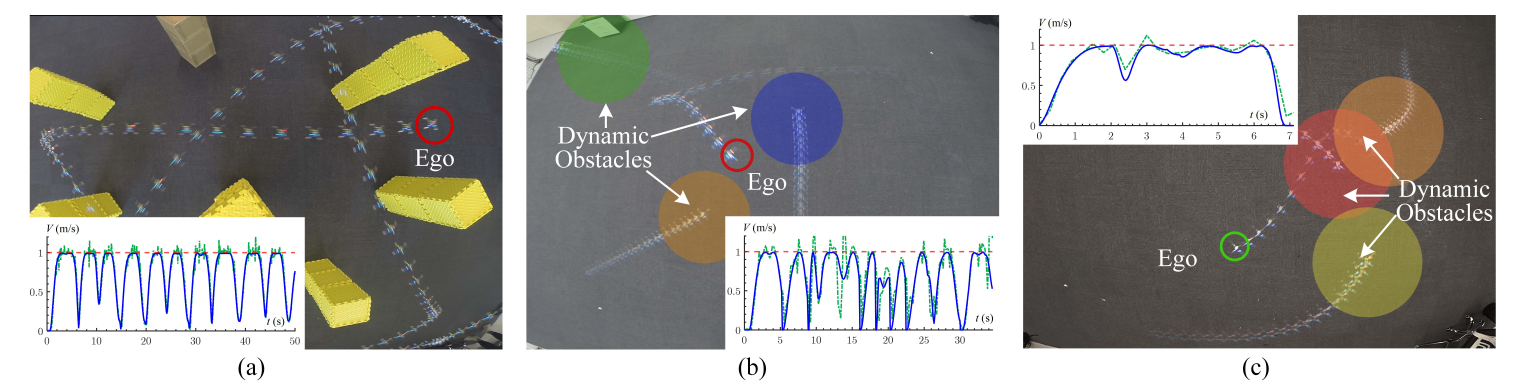} 
	\vspace{-1.0em}
	\caption{Real-world experiment results. The detailed flight processes are provided in the supplementary video.
		(a) Random flight in a static environment.
		(b) Random flight in a dynamic environment.
		(c) Penetration flight in a dynamic environment.}
	\label{fig: real-world experiments}
	\vspace{-1.5em}
\end{figure*}

The real-world experiments are performed to further validate the practicality of TRUST-Planner.
The flight platform is based on the Crazyfile 2.1\footnote{[Online]. \href{https://www.bitcraze.io/products/crazyflie-2-1/}{https://www.bitcraze.io/products/crazyflie-2-1/}.} nanoscale drone, integrated with an OptiTrack motion capture system for 
indoor localization.
Our method is implemented on the ground station as a real-time planning terminal, using Crazyswarm 2 \cite{preiss_crazyswarm_2017} to communicate with the drones via radio link.
The dynamic constraints are $v_{\max} = 1 \, \mathrm{m/s}$, $a_{\max} = 3 \, \mathrm{m/s^2}$.
Three different scenarios are designed: 1) Random flight in a static environment; 2) Random flight in a dynamic environment; 3) Penetration in a dynamic environment.
The results are shown in Fig.\ref{fig: real-world experiments} (see supplementary video \footnote{[Online]. \href{https://www.bilibili.com/video/BV1RJWqzqEMz/}{https://www.bilibili.com/video/BV1RJWqzqEMz/}. Part 2-4.} for details).
During the flight, the ego continuously replans its trajectories to avoid obstacles while heading towards assigned goals.
Especially for the dynamic scenarios, since the ego does not know the precise motion of the dynamic obstacles, autonomous flights in such unknown environments are quite challenging.
However, the TRUST-Planner can still realize millisecond-level trajectory replanning and avoid unknown maneuvers of obstacles.
In terms of the speed profiles, the ego tends to maintain near maximum speed for quickly traversing the obstacle area.
In summary, the real-world experiments illustrate the excellent computational efficiency and applicability of the proposed method.

\section{Conclusions and Future Work}

This paper presents a novel spatial-temporal topological trajectory planning framework, namely TRUST-Planner for AAVs' operating in complex dynamic environments.
It integrates a DEV-PRM frontend for efficient topological path exploration, a UTF-MINCO backend for fast trajectory optimization,
and an incremental multi-branch trajectory management framework to maintain topological diversity and enable trajectory decision.
Extensive simulations and real-world experiments demonstrate that TRUST-Planner achieves superior performance in terms of robustness, trajectory safety and efficiency.
Currently, TRUST-Planner relies on the prediction of dynamic obstacles' motion, which is based on 3D position and velocity data provided by a motion capture system.
In future work, we aim to extend the framework to more practical scenarios by developing obstacle identification and prediction based on onboard sensing, such as vision or LiDAR, to further enhance the applicability and autonomy of our method.


\bibliographystyle{IEEEtranTIE}
\bibliography{IEEEabrv,Reference} 

\vspace{-3em}
\begin{IEEEbiography}
	[{\includegraphics[width=1in,height=1.25in,clip,keepaspectratio]{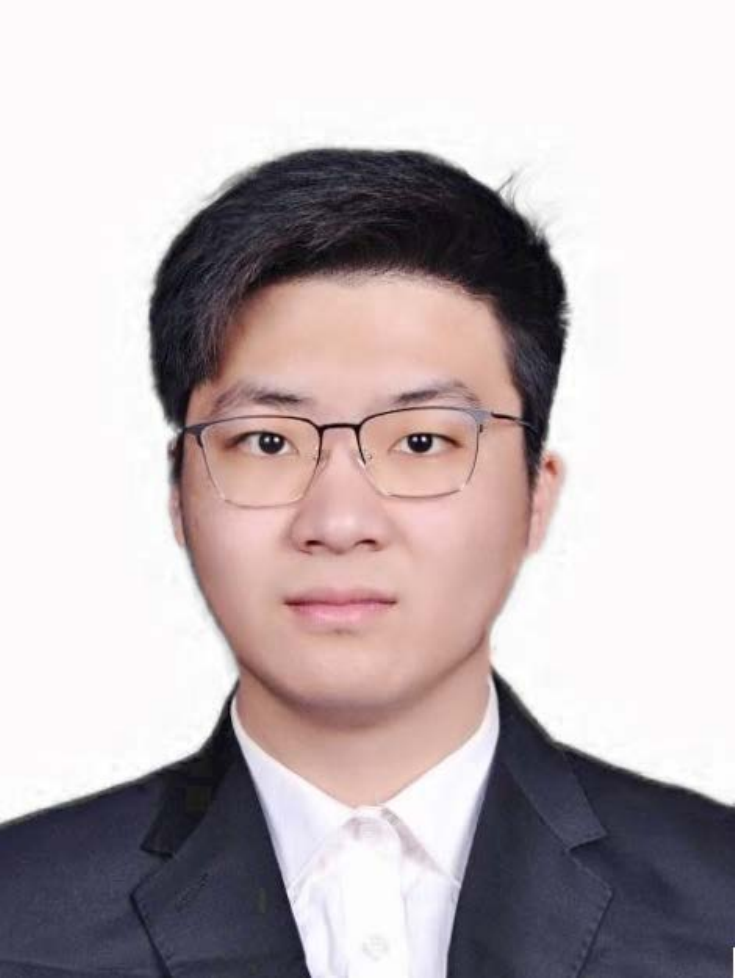}}]{Junzhi Li}
	received the B.S. degree in flight vehicle design and engineering from the School of Aerospace Engineering, Beijing Institute of Technology, Beijing, China, in 2020. He is currently working toward a Ph.D. degree in aeronautical and astronautical science and technology with Beijing Institute of Technology.

	His research interests include trajectory optimization and control of unmanned flight vehicles.
\end{IEEEbiography}
\vspace{-3em}
\begin{IEEEbiography}
	[{\includegraphics[width=1in,height=1.25in,clip,keepaspectratio]{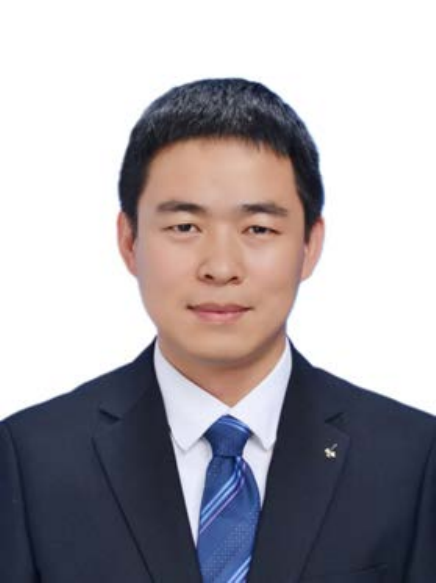}}]{Jingliang Sun}
	received the B.S. degree in automation from the Tianjin University of Technology, China, and the Ph.D. in control theory and control engineering from the Nanjing University of Aeronautics and Astronautics, China.

	He is an associate professor with the School of Aerospace Engineering, Beijing Institute of Technology. His research interests include adaptive dynamic programming, cooperative guidance and control, and mission planning.
\end{IEEEbiography}
\vspace{-3em}
\begin{IEEEbiography}
	[{\includegraphics[width=1in,height=1.25in,clip,keepaspectratio]{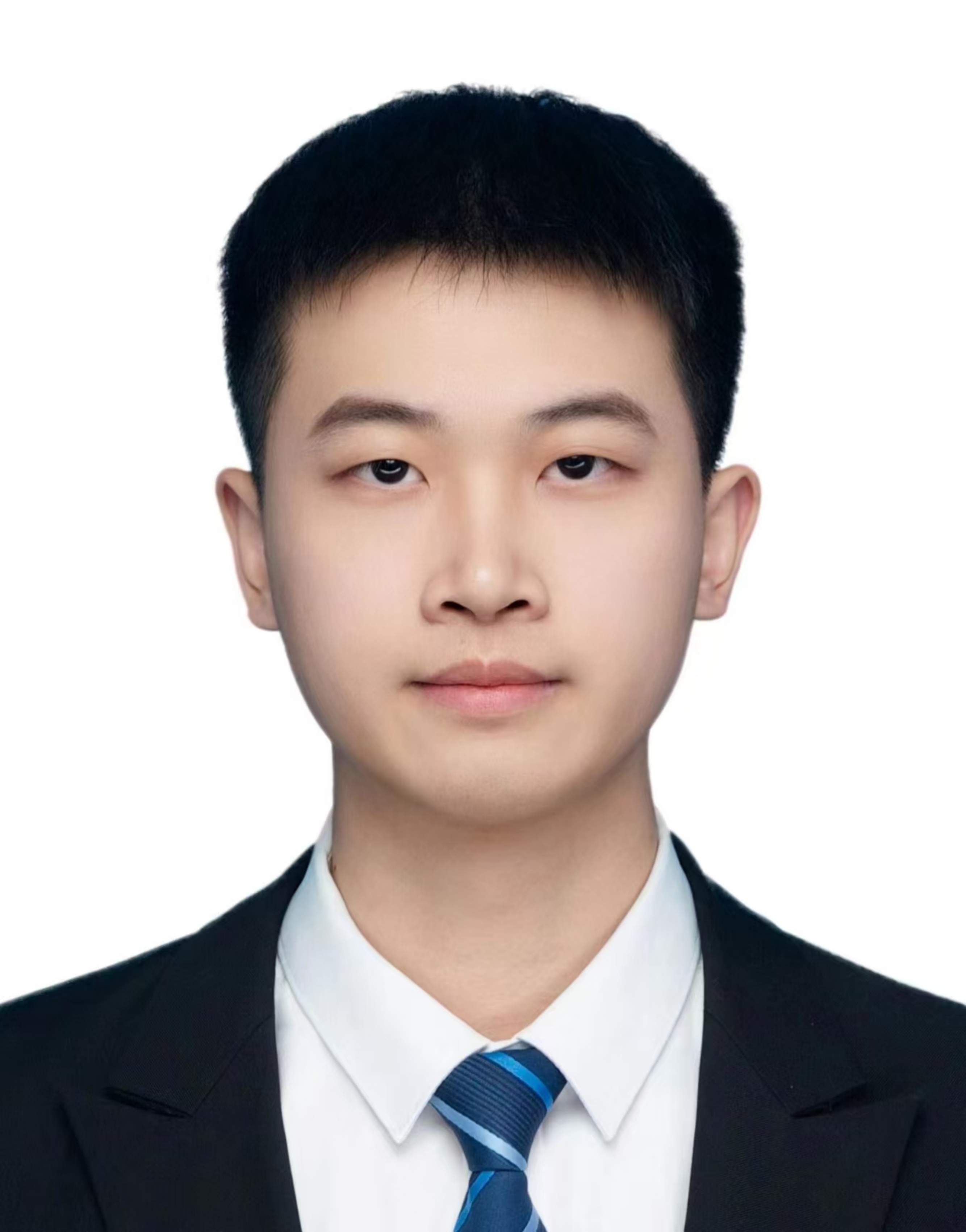}}]{Jianxin Zhong}
	received the B.S. degree in flight vehicle design and engineering from Beijing Institute of Technology, Beijing, China, in 2023. He is currently working toward the M.S. degree in aeronautical and astronautical science and technology with Beijing Institute of Technology.

	His research interests include multi-agent path planning and cooperative task assignment of unmanned aerial vehicles.
\end{IEEEbiography}
\vspace{-3em}
\begin{IEEEbiography}
	[{\includegraphics[width=1in,height=1.25in,clip,keepaspectratio]{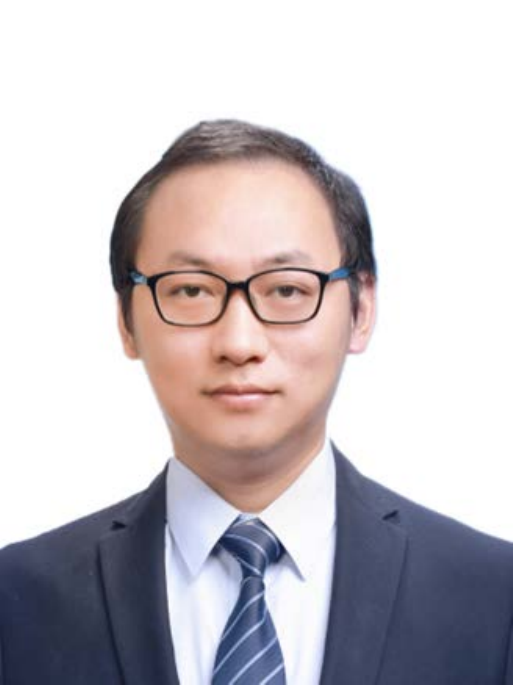}}]{Teng Long}
	received the B.S. degree in flight vehicle propulsion engineering and
	the Ph.D. in aeronautical and astronautical science and technology from
	Beijing Institute of Technology, China, in 2004 and 2009,
	respectively.

	He is currently a Professor and Dean with the School of Aerospace
	Engineering, Beijing Institute of Technology. His research interests
	include multidisciplinary optimization and its applications to
	flight vehicle conceptual design, cooperative mission
	planning and decision-making.
\end{IEEEbiography}

\end{document}